\documentclass{llncs}

\usepackage{times}
\usepackage{epsfig}
\usepackage{graphicx}
\usepackage{amsmath}
\usepackage{amssymb}
\usepackage{subcaption}
\usepackage[utf8]{inputenc}
\usepackage{multicol}
\usepackage[pagebackref=true,breaklinks=true,letterpaper=true,colorlinks,bookmarks=false]{hyperref}

\begin{document}

\title{Automatic skin lesion segmentation on dermoscopic images by the means of superpixel merging}

\author{Diego Patiño\orcidID{0000-0003-4808-8411}\inst{1} \and Jonathan Avendaño\orcidID{0000-0001-7599-0481} \inst{1} \and John W. Branch\orcidID{0000-0002-0378-028X}\inst{1}}

\institute{Universidad Nacional de Colombia, Faculty of Mines, Calle 80 \#65 - 223, Medellín, Colombia \\
	\{dapatinoco, jdavendanoo, jwbranch\}@unal.edu.co}


\maketitle

\begin{abstract}

We present a superpixel-based strategy for segmenting skin lesion on dermoscopic images. The segmentation is carried out by over-segmenting the original image using the SLIC algorithm, and then merge the resulting superpixels into two regions: healthy skin and lesion. The mean RGB color of each superpixel was used as merging criterion. The presented method is capable of dealing with segmentation problems commonly found in dermoscopic images such as hair removal, oil bubbles, changes in illumination, and reflections images without any additional steps. The method was evaluated on the PH2 and ISIC 2017 dataset with results comparable to the state-of-art.
	
\end{abstract}


\section{Introduction}

Melanoma is a type of skin cancer that can occur in any type of skin, and it is the leading cause of death among all skin cancer. A well-known criterion for early-stage melanoma detection is the ``ABCDE'' rule, which was designed for human-based visual analysis of a skin lesion. The ``ABCDE'' rule uses different visual cues to classify the lesion type: (A) Asymmetry, (B) irregularity of the Borders, (C) presence of specific Colors, (D) lesion shape and size, and (E) evaluation of the lesion evolution over time.

Early detection of malignant melanoma in dermoscopy images is crucial and critical since its detection in the early stages can be helpful to cure it. Computer Aided Diagnosis systems are becoming very helpful in facilitating the early detection of cancers for dermatologists. However, some challenges are present in this type of images that make difficult for computers to identify the lesion from healthy skin: presence of hair, changes in illumination, different types of skin, reflections, and oil bubble are some of the most common artifacts found in this type of images.

This paper focuses on constructing an accurate fully-automatic method to identify skin lesions in dermoscopy images to facilitate accurate, fast and more reliable identification and analysis. We used well-stated computer vision algorithms to segment skin lesion by first over-segmenting the original image and then merging the resulting regions into two types: skin lesion and healthy skin.

One fundamental difference of our approach compared to previous work is the fact that our segmentation method does not require a pre-processing step for correcting image artifacts like hair removal or illumination correction in contrast with most approaches in the literature. The stated work outputs a final binary mask containing the image's area where the lesion is located without the need for any user-defined parameter.

The segmentation strategy was tested on the PH$^2$\cite{Mendonc2013} and the ISIC 2017 challenge (part 1) \cite{Codella2017a} datasets. PH$^2$ consists of 200 images divided into three groups: common lesions, atypical nevi, and melanomas. ISIC, instead, has 2000 dermoscopic images with corresponding ground-truth segmentation. The experiments using our approach show comparable results with respect to previous works, achieving sensitivities and accuracy of around 92\% in almost all tests, and demonstrate that our algorithm can be a suitable tool for the development of CAD support systems for the early detection of skin lesions.

The rest of this document is organized as follows: Section \ref{previous_work} reviews previous works on skin lesion segmentation. Section \ref{segmentation} presents a method for skin lesion segmentation based on superpixel approach. Section \ref{results} experimental results are presented. Finally, section \ref{conclusions} concludes the paper and states potential future work.

\section{Related work}
\label{previous_work}

Many segmentation algorithms have been proposed in the literature to deal with the problem of accurately segmenting skin lesion on dermatoscopic images while classifying several types of lesions\cite{Celebi2015,Celebi2010,Abuzaghleh2014,Barata2015}. Such approaches share a common structure that consists of first removing artifacts like hair and oil bubbles in the images, and then segment the skin lesion using different combinations of thresholding, clustering, and morphological operations.

One of the more representatives works on the subject is \cite{Pennisi2016}, where the authors defined a complete framework for skin lesion segmentation and classification based on Delaunay Triangulation. The polygons in such triangulation adapt to the lesion border according to a color criterion. In an early work, Sabbaghi et al. \cite{Sabbaghi2015} also presented an interesting approach for this type of classification. They used QuadTree algorithm to subsequently group pixels with similar color properties and then analyze which color are present in each type of lesion. The authors did not perform any segmentation and do the classification using the whole set of pixels in the images.

Although Deep Learning is considered the state-of-the-art for a vast number of computer vision tasks, only in recent years deep learning methods have started being applied to skin-disease segmentation and classification using different architectures and schemes\cite{Codella2015,Li2017,Gao2017,Codella2017}.

In this paper, we proposed a novel superpixel-based method for segmenting skin lesion on dermoscopic images without the need of pre-processing operations for artifacts removals such as reflections and hair. The main idea of our method is to over-segment the image into several superpixels, and then merge those which belong to the skin lesion to create a binary mask containing the area where the lesion is located. 

\section{Skin lesion segmentation}
\label{segmentation}

In this section, we describe our method for skin lesion segmentation on dermoscopic images. First, we use SLIC algorithm to obtain an over-segmentation of the original dermoscopic image (I). Later, a binary search is performed to find the best threshold to greedily merges all the superpixels into two regions: lesion and healthy skin. Finally, a post-processing stage is run to remove small superpixels and smooth the final segmentation.

\subsection{Data set}

In order to test the proposed approach, we used the PH$^2$ \cite{Mendonc2013} data set released by the Universidade do Porto, in collaboration with the Hospital Pedro Hispano in Matosinhos, Portugal. This data set contains 200 RGB dermoscopic images of melanocytic lesions, including 80 common nevi, 80 atypical nevi, and 40 melanomas. All images have manually generated ground-truth segmentation of the skin lesion of each image.
	
The method was also tested on the ISIC 2007 challenge dataset \cite{Codella2017a} which consist of 2000 dermoscopic images with corresponding manually generated ground-truth.

\subsection{Superpixel segmentation}

Superpixels methods over segment an image by grouping pixels into units called superpixels. One of the most well-known algorithms for superpixel segmentations is Simple Linear Iterative Clustering (SLIC) developed by Achanta et al. in 2012\cite{achanta2012}. In SLIC the pixel grouping is done by clustering pixels with k-means algorithm using as features their color intensities plus their $(i, j)$ coordinates of each pixel, weighted by a factor $\alpha$. The parameter $\alpha$ helps to regularize the trade-off between spatial clustering and color clustering. Additionally, SLIC modifies the basic k-means algorithm applying some constraints to prevent that unconnected pixels could belong to the same superpixel.

For segmenting the skin lesion from the whole image, our method performs a SLIC superpixel segmentation on the RGB dermoscopic images. Ideally, in the over-segmented image, some superpixel's boundary should match partially the boundary of the object that is intended to segment. 

\begin{figure}[ht]
	\centering
	\begin{subfigure}{.300\textwidth}
		\begin{center}
			\includegraphics[width=1\linewidth]{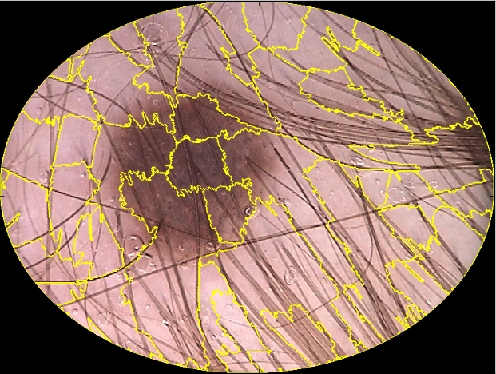}
		\end{center}
	\end{subfigure}
	\begin{subfigure}{.300\textwidth}
		\begin{center}
			\includegraphics[width=1\linewidth]{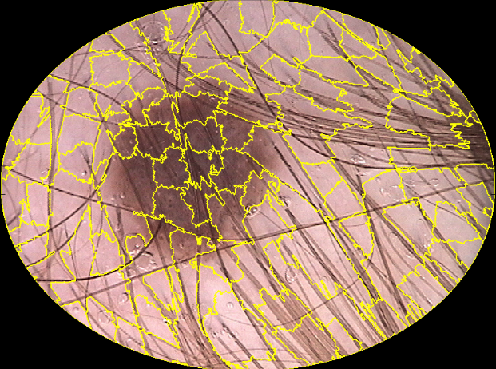}
		\end{center}
	\end{subfigure}

	\begin{subfigure}{.300\textwidth}
		\begin{center}
			\includegraphics[width=1\linewidth]{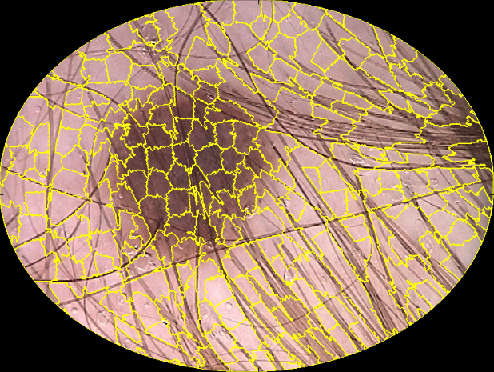}
		\end{center}
	\end{subfigure}
	\begin{subfigure}{.300\textwidth}
		\begin{center}
			\includegraphics[width=1\linewidth]{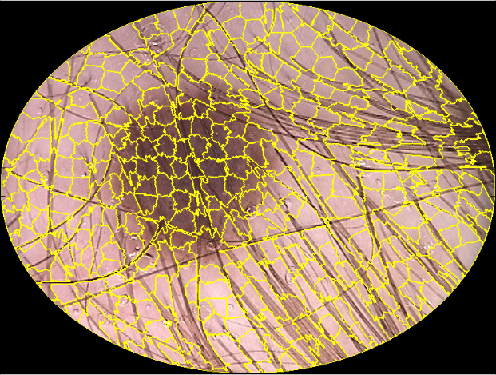}
		\end{center}
	\end{subfigure}
	\caption{SLIC segmentation for several $k$ values. Left to right and top to bottom: $k=100, 200, 400, 600$.}
	\label{fig:seg_k_values}
\end{figure}

SLIC requires a parameter $k$ indicating the desired number of superpixels in the resulting image. A low value of $k$, although faster, leads to few and bigger superpixels where the superpixel's boundary does not entirely match the boundary of the object. With higher values of $k$, too much over-segmentation is done resulting in longer execution times with no significant improvement in the resulting segmentation. Figure \ref{fig:seg_k_values} shows the resulting superpixel segmentation for two images and four different values of $k$. In our experiments we empirically set $k=400$.

All original images have a dark shadow around the four corners of the images. This is an effect of the illumination setup in the dermatoscope device that was used to capture the images. To remove such effect, a binary mask is applied to the resulting SLIC segmentation. The mask is defined as the maximum ellipse inscribed in the image area. 


\subsection{Superpixel merging}

SLIC segmentation produces an over-segmented image with approximately $k$ pieces. Hence, a merging operation must be done to separate the skin lesion (foreground superpixels) from the region containing pixels with no lesion (background superpixels). SLIC's output can be seen as a label image (L) where each integer label corresponds to one superpixel. From this representation, a Region Adjacency Graph (RAG) is constructed. Each node is equipped with a list of properties derived from its RGB intensity values: Mean color, total color, and pixel count. 

Superpixel merging is then done by greedily combining pairs of nodes using the euclidean distance of the mean color of each node as criterion. If this difference is less than a threshold $t$, then the two nodes will merge, and a new node is created. The new node's properties are then calculated based on the merging nodes.

A binary search determines the optimal threshold by running the merging procedure with several values of $t$. Assuming RGB images with 3 channels, and intensity levels of each pixel in the range: $p \in [0, 255]$, the distance between the mean color of two different superpixels will in the interval $(0, 500)$. Therefore, the binary search will be constrained to such interval. The search converges when for a given $t$ only two superpixels remain, or in the case that the maximum number of iterations is reached, or if $\Delta t <  \epsilon$. To avoid trivial results with just one region, If the procedure result has only one superpixel, then it returns the last known threshold with at least two regions.

\subsection{Post-processing}
\label{post-processing}

After merging with the optimal threshold, a post-processing step is performed to determine the final segmentation and smooth the results. First, All remaining regions with an area lower than $2\%$ from the total area of the image are removed. Small superpixels remaining after merging are considered as noise because the two expected regions should be relatively significant areas corresponding to the lesion or healthy skin.

The merging produces a label image (L) where each superpixel is associated to an integer value. At this point, it is not possible to determine whether a label corresponds to background or skin lesion. Therefore, image O is created by applying adaptive equalization to a gray-scale version of the original image and subsequently segmenting it using an Otsu threshold. Each label in L is compared with O by calculating the Jaccard similarity index between the two areas. The label with the maximum Jaccard index is selected as the final segmented image after applying binary fill holes, and a morphological dilation with a disk-shaped structural element of radius $8$.

\section{Experimental results}
\label{results}

To compare our segmentation with other methods, we performed the same set of test presented in \cite{Pennisi2016}. Hence, we used PH$^2$ ground-truth lesion segmentation to evaluate the outcome of our algorithm using four metrics: sensitivity, specificity, accuracy, and F-measure. Additionally, the data set was split into four subsets: all images, common lesions, atypical nevi, and melanomas. Tables \ref{table:all} to \ref{table:melanoma} show the segmentation results as presented in \cite{Pennisi2016}, but extended with our results.


In each of the four tests, the presented strategy achieved significantly better sensitivity than previous works. This means that pixels belonging to a skin lesion are segmented correctly in a higher proportion compared with other methods (fewer false positives). A similar situation occurs with the F-measure metric. Also, better accuracy was also obtained in all test cases except with melanoma lesions.

However, specificity did not perform as well as the other three metrics. In all cases, specificity was slightly less than the highest value. This means our method tends to classify healthy pixels as lesion over-estimating the lesion area.

\begin{table}[ht]
\caption{Segmentation results on the PH2 dataset.}
\begin{subtable}{0.48\textwidth}
	\caption{Using all 200 images (melanocytic nevi, dysplasic nevi, malignant lesions).}
	\label{table:all}
	\begin{center}
		\scriptsize
		\begin{tabular}{p{1.0cm}p{.8cm}p{.8cm}p{2.0cm}p{1.1cm}}
			\hline\noalign{\smallskip}
			Method & Sens. & Spec. & Acc. & F-measure \\ 
			\noalign{\smallskip}
			\hline
			\noalign{\smallskip}
			JSEG\cite{JSGE2001} & 0.7108 & 0.9714 & 0.8947 $\pm\, 0.0176$ & 0.7554 \\
			SRM\cite{SRM2010} & 0.1035 & 0.8757 & 0.6766 $\pm\, 0.0346$ & 0.1218 \\
			KPP & 0.4147 & 0.9581 & 0.7815 $\pm\, 0.0356$ & 0.5457 \\
			K-means & 0.7291 & 0.8430 & 0.8249 $\pm\, 0.0107$ & 0.6677 \\
			Otsu & 0.5221 & 0.7064 & 0.6518 $\pm\, 0.0203$ & 0.4293 \\
			Level Set\cite{levelset} & 0.7188 & 0.8003 & 0.7842 $\pm\, 0.0295$ & 0.6456 \\
			ASLM\cite{Pennisi2016} & 0.8024 & \textbf{0.9722} & 0.8966 $\pm\, 0.0276$ & 0.8257 \\
			Our method & \textbf{0.9104} & 0.8973 & \textbf{0.9039 $\mathbf{\pm\, 0.1419}$} & \textbf{0.8918} \\
			\hline
		\end{tabular}
	\end{center}
\end{subtable}
\hspace{0.3cm}
\begin{subtable}{0.48\textwidth}
	\caption{Only 80 melanocytic nevi images (common healthy lesions).}
	\label{table:common}
	\begin{center}
		\scriptsize
		\begin{tabular}{p{1.0cm}p{.8cm}p{.8cm}p{2.0cm}p{1.1cm}}
			\hline\noalign{\smallskip}
			Method & Sens. & Spec. & Acc. & F-measure \\ 
			\noalign{\smallskip}
			\hline
			\noalign{\smallskip}
			JSEG\cite{JSGE2001} & 0.6977 & \textbf{0.9783} & 0.9370 $\pm\, 0.0027$ & 0.7265 \\
			SRM\cite{SRM2010} & 0.0751 & 0.9332 & 0.7250 $\pm\, 0.0277$ & 0.0611 \\
			KPP & 0.3360 & 0.9566 & 0.7912 $\pm\, 0.0241$ & 0.3960 \\
			K-means & 0.7008 & 0.8767 & 0.8466 $\pm\, 0.8467$ & 0.6004 \\
			Otsu & 0.4777 & 0.7832 & 0.6911 $\pm\, 0.0193$ & 0.3658 \\
			Level Set\cite{levelset} & 0.7069 & 0.8262 & 0.7996 $\pm\, 0.0264$ & 0.5856 \\
			ASLM\cite{Pennisi2016} & 0.8717 & 0.9760 & 0.9477 $\pm\, 0.0032$ & 0.8690 \\
			Our method & \textbf{0.9212} & 0.9642 & \textbf{0.9524 $\mathbf{\pm\, 0.0637}$} & \textbf{0.9292} \\
			\hline
		\end{tabular}
	\end{center}
\end{subtable}

\begin{subtable}{0.48\textwidth}
	\caption{Only 80 dysplasic nevi images (atypical moles).}
	\label{table:atypical}
	\begin{center}
		\scriptsize
		\begin{tabular}{p{1.0cm}p{.8cm}p{.8cm}p{2.0cm}p{1.1cm}}
			\hline\noalign{\smallskip}
			Method & Sens. & Spec. & Acc. & F-measure \\ 
			\noalign{\smallskip}
			\hline
			\noalign{\smallskip}
			JSEG\cite{JSGE2001} & 0.7435 & 0.9708 & 0.9236 $\pm\, 0.0065$ & 0.7768 \\
			SRM\cite{SRM2010} & 0.1042 & 0.8954 & 0.6812 $\pm\, 0.0358$ & 0.0919 \\
			KPP & 0.2895 & 0.9446 & 0.7512 $\pm\, 0.0261$ & 0.3568 \\
			K-means & 0.7650 & 0.8804 & 0.8501 $\pm\, 0.0065$ & 0.6914 \\
			Otsu & 0.5515 & 0.7579 & 0.6779 $\pm\, 0.0193$ & 0.4372 \\
			Level Set\cite{levelset} & 0.7364 & 0.8237 & 0.7985 $\pm\, 0.0346$ & 0.6532 \\
			ASLM\cite{Pennisi2016} & 0.8640 & \textbf{0.9733} & 0.9271 $\pm\, 0.0099$ & 0.8689 \\
			Our method & \textbf{0.9225} & 0.9354 & \textbf{0.9314 $\mathbf{\pm\, 0.0841}$} & \textbf{0.9112} \\
			\hline
		\end{tabular}
	\end{center}
\end{subtable}
\hspace{0.3cm}
\begin{subtable}{0.48\textwidth}
	\caption{Only 40 melanoma images (malignant lesions).}
	\label{table:melanoma}
	\begin{center}
		\scriptsize
		\begin{tabular}{p{1.0cm}p{.8cm}p{.8cm}p{2.0cm}p{1.1cm}}
			\hline\noalign{\smallskip}
			Method & Sens. & Spec. & Acc. & F-measure \\ 
			\noalign{\smallskip}
			\hline
			\noalign{\smallskip}
			JSEG\cite{JSGE2001} & 0.6746 & 0.9593 & \textbf{0.7591 $\mathbf{\pm\, 0.0456}$} & 0.7710 \\
			SRM\cite{SRM2010} & 0.2234 & 0.7512 & 0.4148 $\pm\, 0.0366$ & 0.2852 \\
			KPP & 0.2648 & 0.7623 & 0.4324 $\pm\, 0.0336$ & 0.3589 \\
			K-means & 0.5971 & 0.4870 & 0.5524 $\pm\, 0.0211$ & 0.6064 \\
			Otsu & 0.7073 & 0.7015 & 0.7249 $\pm\, 0.0214$ & 0.7503 \\
			Level Set\cite{levelset} & 0.7141 & 0.7010 & 0.7313 $\pm\, 0.0230$ & 0.7550 \\
			ASLM\cite{Pennisi2016} & 0.5404 & \textbf{0.9597} & 0.6615 $\pm\, 0.0506$ & 0.6524 \\
			Our method & \textbf{0.8645} & 0.6870 & 0.7519 $\pm\, 0.2216$ & \textbf{0.7779} \\
			\hline
		\end{tabular}
	\end{center}
\end{subtable}
\end{table}

\begin{figure}[ht]
	\centering
	\begin{subfigure}{.245\textwidth}
		\begin{center}
			\includegraphics[width=1\linewidth]{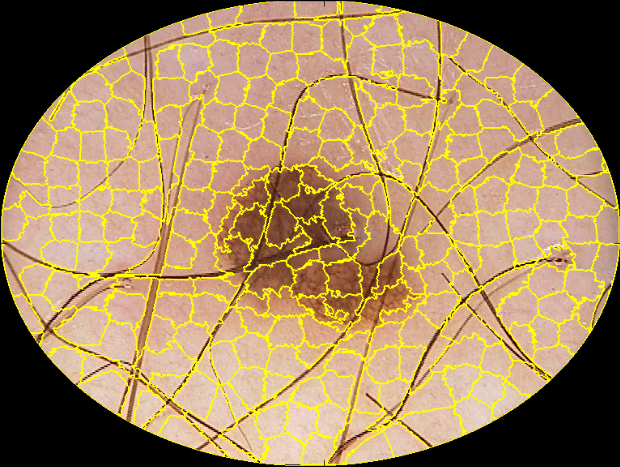}
		\end{center}
	\end{subfigure}
	\begin{subfigure}{.245\textwidth}
		\begin{center}
			\includegraphics[width=1\linewidth]{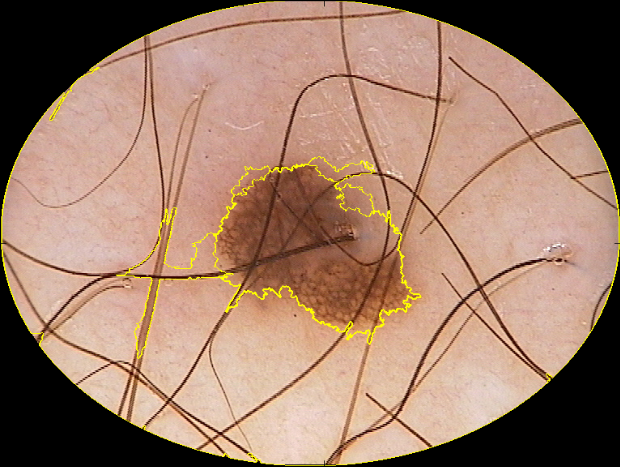}
		\end{center}
	\end{subfigure}
	\begin{subfigure}{.245\textwidth}
		\begin{center}
			\includegraphics[width=1\linewidth]{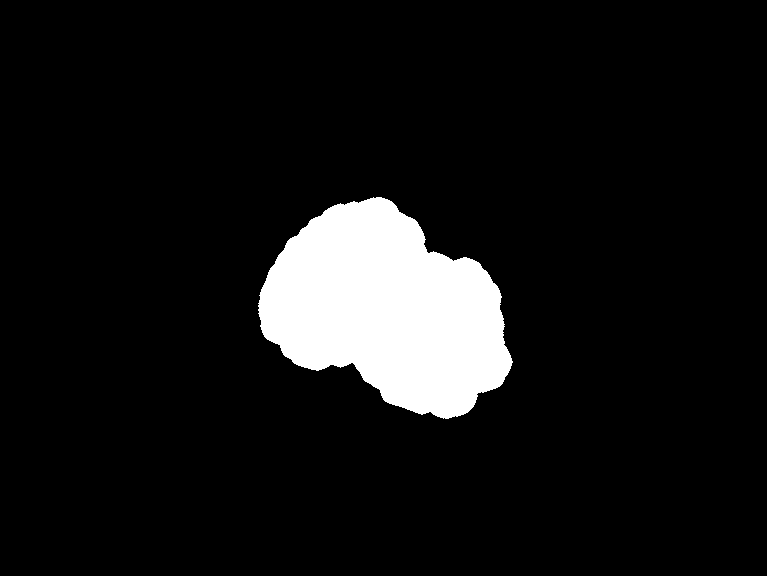}
		\end{center}
	\end{subfigure}
	\begin{subfigure}{.245\textwidth}
		\begin{center}
			\includegraphics[width=1\linewidth]{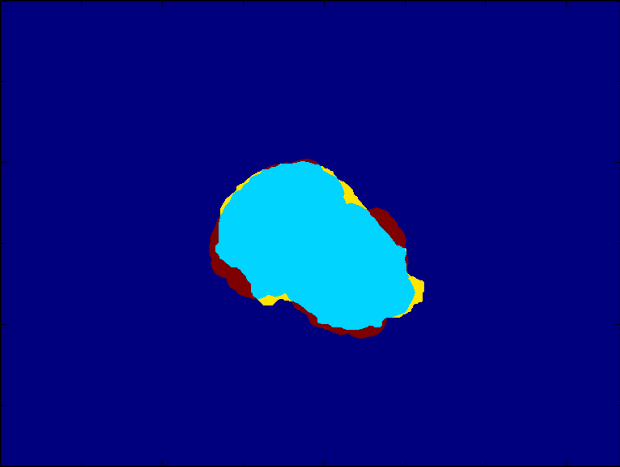}
		\end{center}
	\end{subfigure}
	\begin{subfigure}{.245\textwidth}
		\begin{center}
			\includegraphics[width=1\linewidth]{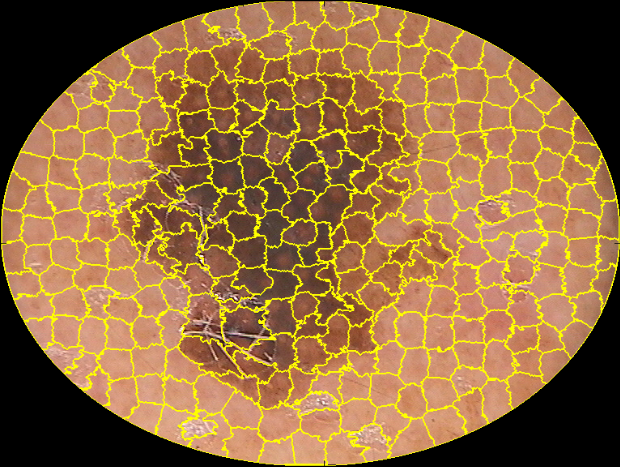}
		\end{center}
	\end{subfigure}
	\begin{subfigure}{.245\textwidth}
		\begin{center}
			\includegraphics[width=1\linewidth]{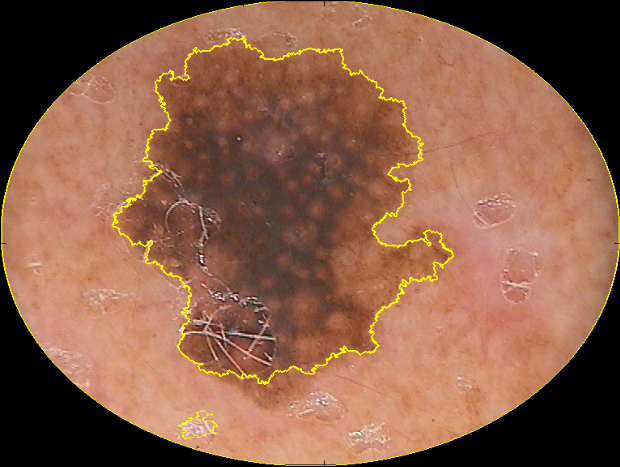}
		\end{center}
	\end{subfigure}
	\begin{subfigure}{.245\textwidth}
		\begin{center}
			\includegraphics[width=1\linewidth]{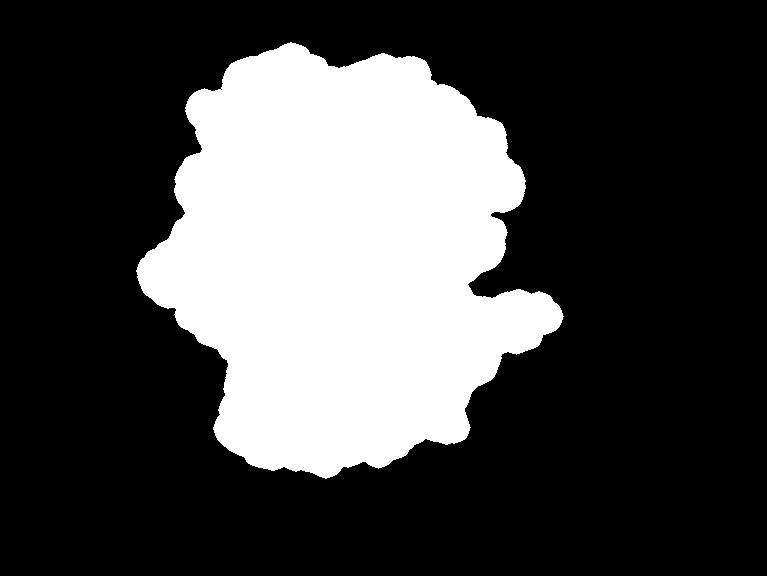}
		\end{center}
	\end{subfigure}
	\begin{subfigure}{.245\textwidth}
		\begin{center}
			\includegraphics[width=1\linewidth]{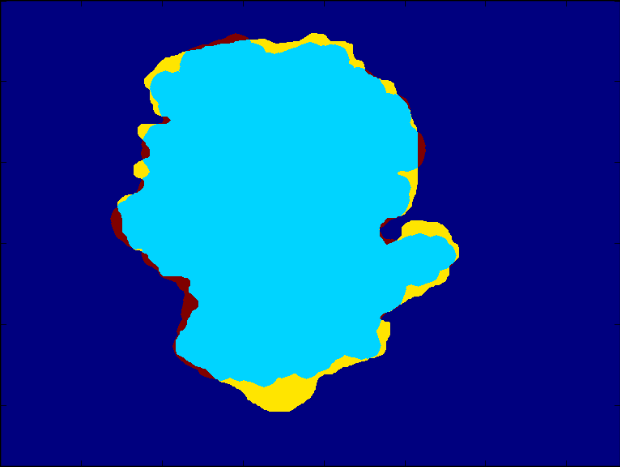}
		\end{center}
	\end{subfigure}
	\begin{subfigure}{.245\textwidth}
		\begin{center}
			\includegraphics[width=1\linewidth]{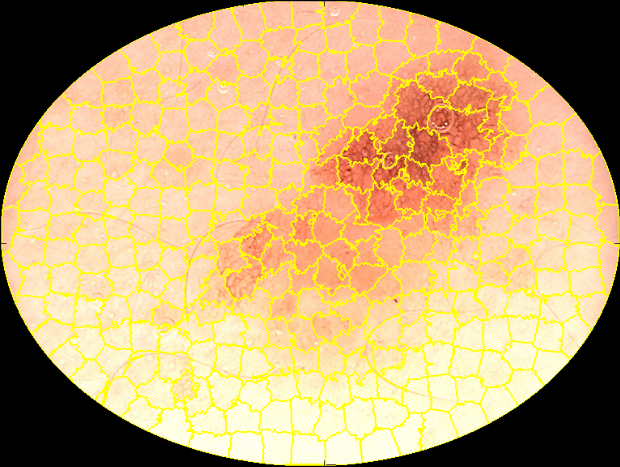}
		\end{center}
	\end{subfigure}
	\begin{subfigure}{.245\textwidth}
		\begin{center}
			\includegraphics[width=1\linewidth]{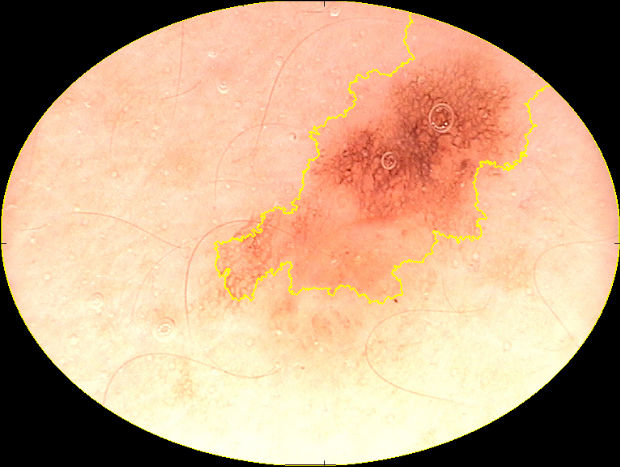}
		\end{center}
	\end{subfigure}
	\begin{subfigure}{.245\textwidth}
		\begin{center}
			\includegraphics[width=1\linewidth]{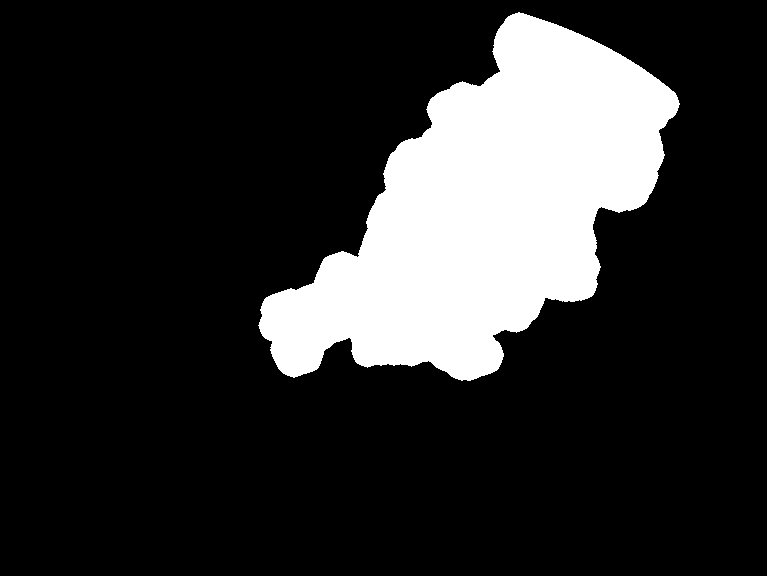}
		\end{center}
	\end{subfigure}
	\begin{subfigure}{.245\textwidth}
		\begin{center}
			\includegraphics[width=1\linewidth]{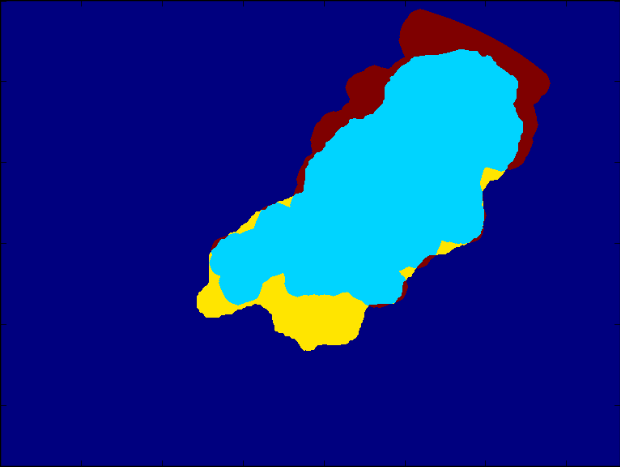}
		\end{center}
	\end{subfigure}
	\caption{Segmentation results for three different images. From left to right: Original image with SLIC segmentation ($k=400$), Merged superpixels, final segmentation binary mask, and ground-truth comparison. TP=light blue, TN=dark blue, FN: yellow and FP: red.}
	\label{fig:seg_results}
\end{figure}

In previous works, hair removal is in many cases addressed with an extra preprocessing step before segmenting the lesion. This step involves the applications of different filters such as directional Gaussian filters\cite{Abuzaghleh2015} to the original image. In contrast, our strategy does not require any additional step since SLIC's superpixels fit the borders of the lesion avoiding hair areas. Figure \ref{fig:seg_results} shows the segmentation results for three different images. Rows one and three correspond to Atypical nevi lesion, and row two is a melanocytic nevus. 

Moreover, experimental results also show that the performance of our method, although slightly lower, is comparable with those achieved by deep learning methods (as shown in table \ref{table:deep_learning}), but without the use of artifacts like fine-tuning or data augmentation.

\begin{table}[ht]
	\caption{Segmentation results compared with the state-of-the deep learning\cite{Codella2017}.}
	\label{table:deep_learning}
	\begin{center}
		\scriptsize
		\begin{tabular}{lcc}
			\hline\noalign{\smallskip}
			Method & Jaccard & Accuracy \\ 
			\noalign{\smallskip}
			\hline
			\noalign{\smallskip}
			Optimized Single & 0.836 & 0.949 \\
			Default Augmentation & 0.828 & 0.947 \\
			No Noise or Dropout & 0.812 & 0.941 \\
			Ensemble of 10 U-Nets  & 0.841 & 0.951 \\
			State-of-art & \textbf{0.843} & \textbf{0.953} \\
			Human Expert Average Agreement & 0.786 & 0.909 \\
			\textbf{our method} & 0.606 & 0.869 \\
			\hline
		\end{tabular}
	\end{center}
\end{table}

\begin{figure}[ht]
	\centering
	\begin{subfigure}{.435\textwidth}
		\begin{center}
			\includegraphics[width=1\linewidth]{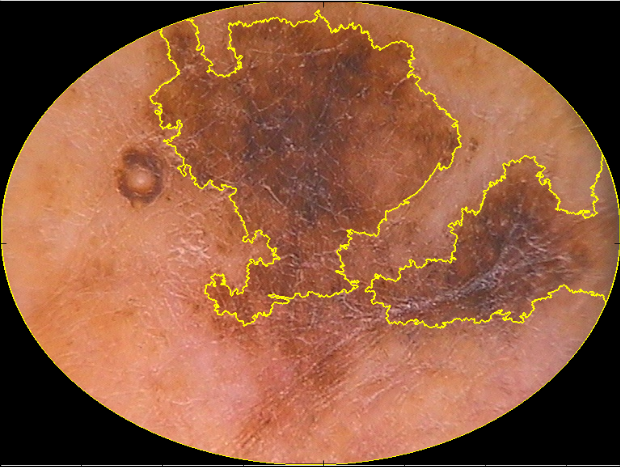}
		\end{center}
	\end{subfigure}
	\begin{subfigure}{.435\textwidth}
		\begin{center}
			\includegraphics[width=1\linewidth]{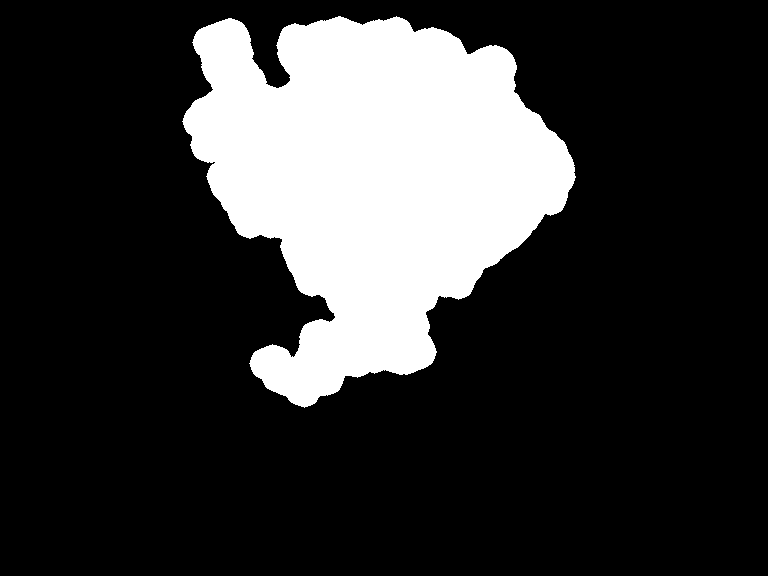}
		\end{center}
	\end{subfigure}
	\caption{Skin lesion with two unconnected regions. The segmentation algorithm only identified one of two parts.}
	\label{fig:two_regions}
\end{figure}

Although the algorithm does an acceptable job segmenting the images, a particular case occurs when the lesion consists of several unconnected regions. In this case, the algorithm only chooses one region as the final segmentation as stated in section \ref{post-processing}, resulting in an incorrect binary mask (See figure \ref{fig:two_regions}).

The source code implementation of the presented method is publicly available at \url{https://github.com/dipaco/mole-classification}, and is entirely written in python 2.7. The implementation uses scipy and scikit libraries for its functionality. All tests ran on a desktop machine with an Intel Core i7 processor of 2,5 GHz.

\section{Conclusions and future work}
\label{conclusions}

In this paper, we proposed a new fully-automatic strategy to segment skin lesion on dermoscopic images. Our method uses the SLIC algorithm to over-segment the image, and then merge the resulting superpixels to produce two regions: healthy skin and skin lesion. The merging criterion is based on the mean color intensity of each superpixel. Our method does not require any form of pre-processing of the original image before segmentation and can produce slightly better results than other approaches. In contrast with other works, our approach is able to deal with the presence of hair in the original image without any additional steps. 

From table \ref{table:all} to \ref{table:melanoma} is possible to conclude that, for most methods, segmentation performance was low in cases where the images contained malignant lesions compared with the case of common and atypical nevi. Despite this, our approach achieved better sensitivity and F-measure, and the second-best accuracy on melanocytic images.


Future work will focus on adjusting the merging criterion to deal with cases where the lesion has more that one unconnected region. 

{\small
\bibliographystyle{splncs}

}

\end{document}